\documentclass[10pt,twocolumn,letterpaper]{article}

\usepackage{iccv}
\usepackage{times}
\usepackage{epsfig}
\usepackage{graphicx}
\usepackage{amsmath}
\usepackage{amssymb}
\usepackage[utf8]{inputenc}
\usepackage{caption}
\usepackage{subcaption}
\usepackage{amsfonts}
\usepackage[cal=rsfso,calscaled=.96]{mathalfa}
\usepackage{mathrsfs}  
\usepackage{txfonts} 
\usepackage{enumerate}
 
\usepackage{amsmath}
\usepackage{algorithm}
\usepackage[noend]{algpseudocode}
\usepackage{times} % assumes new font selection scheme installed
\usepackage{amssymb,longtable,calc}
\usepackage{mathptmx}       
\usepackage[textwidth=2cm,colorinlistoftodos]{todonotes}
\usepackage{xspace}
\usepackage{latexsym}
\usepackage{amsmath,amssymb,graphicx,xspace}
\usepackage{times}   % assumes new font selection scheme installed
 
\usepackage{color,comment,rotating,url,xspace} % author packages
\usepackage{tikz}
\usetikzlibrary{arrows,shadows,shapes,backgrounds,decorations,snakes,fit}
\usepackage{sidecap}
\usepackage{subcaption}
\makeatletter
\usepackage{tikz,float}
\usepackage{tabularx}
\usepackage{xparse}
\usetikzlibrary{shapes,backgrounds,calc}
\tikzset{circle split part fill/.style  args={#1,#2}{%
 alias=tmp@name, % Jake's idea !!
  postaction={%
    insert path={
     \pgfextra{% 
     \pgfpointdiff{\pgfpointanchor{\pgf@node@name}{center}}%
                  {\pgfpointanchor{\pgf@node@name}{east}}%            
     \pgfmathsetmacro\insiderad{\pgf@x}
      \fill[#1] (\pgf@node@name.base) ([xshift=-\pgflinewidth]\pgf@node@name.north) arc
                          (90:270:\insiderad-\pgflinewidth)--cycle;
      \fill[#2] (\pgf@node@name.base) ([xshift=\pgflinewidth]\pgf@node@name.south)  arc
                           (0: 0:\insiderad-\pgflinewidth)--cycle;            %  \end{scope}   
         }}}}}  
 \makeatother

 \newcommand{\bmxt}{\mathbf{x_t}}

\newcommand{\bmzt}{\mathbf{z_t}}

\newcommand{\bmhlt}{h_{t-1}} 

\newcommand{\bmyt}{\mathbf{y_t}}

\newcommand{\pem}{\widetilde{p}} 

\newcommand{\bmct}{\mathbf{c_t}}

\iccvfinalcopy % *** Uncomment this line for the final submission

\begin{document}

%%%%%%%%% TITLE
\title{ A Probabilistic Semi-Supervised Approach to Multi-Task Human Activity Modeling }

\author{Judith B\"utepage \hspace{0.5cm} Hedvig Kjellstr\"om \hspace{0.5cm} Danica Kragic\\
Robotics, Perception and Learning \\
KTH Royal Institute of Technology \\
{\tt\small butepage@kth.se, hedvig@kth.se, dani@kth.se }
 }
 
% \author{First Author\\
%Institution1\\
%Institution1 address\\
%{\tt\small firstauthor@i1.org}
% For a paper whose authors are all at the same institution,
% omit the following lines up until the closing ``}''.
% Additional authors and addresses can be added with ``\and'',
% just like the second author.
% To save space, use either the email address or home page, not both
%\and
%Second Author\\
%Institution2\\
%First line of institution2 address\\
%{\tt\small secondauthor@i2.org}
%}

\maketitle
%\thispagestyle{empty}

%%%%%%%%% ABSTRACT
\begin{abstract}
  Human behavior is a continuous stochastic spatio-temporal process which is governed by semantic actions and affordances as well as latent factors. Therefore, video-based human activity modeling is concerned with a number of tasks such as inferring current and future semantic labels, predicting future continuous observations as well as imagining possible future label and feature sequences. In this paper we present a semi-supervised probabilistic deep latent variable model that can represent both discrete labels and continuous observations as well as latent dynamics over time. This allows the model to solve several tasks at once without explicit fine-tuning. We focus here on the tasks of action classification, detection, prediction and anticipation as well as motion prediction and synthesis based on 3D human activity data recorded with Kinect. We further extend the model to capture hierarchical label structure and to model the dependencies between multiple entities, such as a human and objects. Our experiments demonstrate that our principled approach to human activity modeling can be used to detect current and anticipate future semantic labels and to predict and synthesize future label and feature sequences. When comparing our model to state-of-the-art approaches, which are specifically designed for e.g. action classification, we find that our probabilistic formulation outperforms or is comparable to these task specific models.
\end{abstract}

\let\thefootnote\relax\footnotetext{This work was supported by the EU through the project socSMCs (H2020-FETPROACT-2014) and the Swedish Foundation for Strategic Research. 
}

% --------------------------------------------------------------
% Introduction
% --------------------------------------------------------------
\section{Introduction}
\label{sec:intro}

% initial example
Human behavior is determined by many factors such as intention, need, belief and environmental aspects. For example, when standing at a red traffic light, a person might wait or walk depending on whether there is a car approaching, a police car is parked next to the light or they are looking on their mobile phone. This poses a problem for computer vision systems as they often only have access to a visual signal such as single view image sequences or 3D skeletal recordings. Based on this signal, the current class label or subsequent labels and trajectories need to be determined. In the following discussion we focus on action labels, but the class labels can also describe other factors, e.g. environmental aspects such as affordances or object identity. 
% ---------------------------------------------------------------
% init figure
\begin{figure}[t]
\centering
\includegraphics[width=0.46\textwidth]{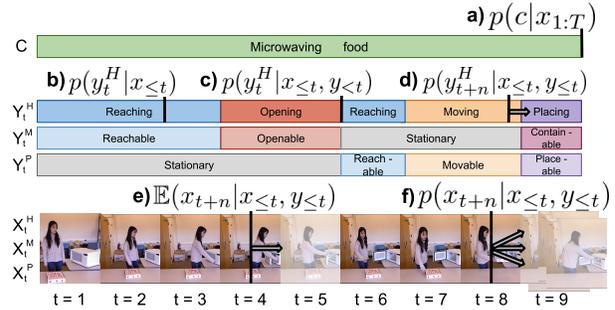}
\caption{ Among others, human activity modeling is concerned with \textbf{a)} action classification, \textbf{b)} action prediction, \textbf{c)} action detection, \textbf{d)} action anticipation, \textbf{e)} motion prediction and \textbf{f)} motion synthesis. The black bars indicate when the respective decision, e.g. classification, is made. Images belong to the CAD - 120 dataset \cite{koppula2013learning}.} \label{fig:init} \vspace{-1.5em}
\end{figure}

% human activity modeling
% The field of human activity modeling can be divided into several tasks, some of which are listed in Table \ref{tab:actions} and visualized in Figure \ref{fig:init}, in which we identify classification, prediction, detection and anticipation of semantic labels as well as prediction and synthesis of continuous human motion. 

% ---------------------------------------------------------------
% table tasks
\begin{table*}[t!]   
\caption{\label{tab:actions} Comparison of data types and tasks for different methods concerned with human activity modeling.  }\vspace{-0.6cm}
\center
\begin{tabular}{|l|l|l|l| }
\hline 
 \multicolumn{1}{|c|}{Method} &  \multicolumn{1}{|c|}{Training data} &  \multicolumn{1}{|c|}{Testing input data}  &  \multicolumn{1}{|c|}{Task}\\ \hline
Action classification &   segmented sequence, labels & segmented sequence  & classify at the end of sequence \\ \hline
Action prediction &   segmented sequence, labels &  segmented sequence & classify as early as possible \\ \hline
Action detection &   segmented sequence, labels &  sequence & detect action onset and classify \\ \hline
Action anticipation &   segmented sequence, labels &  sequence up to time $t$ & predict actions after $t$ \\ \hline
Motion prediction &    sequence, (labels)  &  sequence, (labels) up to $t$ & predict sequence after $t$ \\ \hline
Motion synthesis &    sequence, (labels)  &  sequence, (labels) up to $t$ & generate different sequences after $t$ \\ \hline
\end{tabular}
 \vspace{-1.0em}
\end{table*}

% discrete problems
Most approaches towards human activity modeling focus on problems either concerned with discrete, semantic labels or on continuous trajectory prediction as listed in Table \ref{tab:actions}. Label classification, prediction and detection (Figure \ref{fig:init}a), b) and c)) are supposed to classify observed trajectories either at the end of a sequence (classification), as soon as possible (prediction) or at action onset (detection). Only action anticipation (Figure \ref{fig:init}d)) is concerned with inferring labels of future actions. Human motion prediction and synthesis (Figure \ref{fig:init}e) and \ref{fig:init}f)) on the other hand aim at modeling the future continuous motion trajectories given past observations. Compared to prediction, motion synthesis should anticipate different possible trajectories instead of only the most likely one.

From a modeling perspective, these different types of tasks and mixed categorical and continuous data should influence each other. A model that is able to anticipate a future label should be better at detecting the actual onset of the action. If a model knew that an observed human wants to drink from a nearby glass, the space of possible trajectories would be highly constrained to reaching movements. Likewise, if a model had predicted a reaching trajectory, the inference of future semantic labels would rank "lifting" more likely than "walking".
% what do we require
However, most developed models make not use of this symbiosis and solve only one of the problems in Table \ref{tab:actions} (as discussed in the related work Section \ref{sec:related}). Additionally, in applications that require all tasks to be solved simultaneously, such as in real-time human-robot interaction, this task division requires the deployment of several heavy deep learning architectures which is unfeasible with low-end equipment.

In order to solve these problems simultaneously with a single model, we require a generative model that can represent complex spatio-temporal patterns based on noisy data recordings. It should be able to model the feature space and the label space over time, even if meta-data and meta-labels or hierarchical label structures are present. If these prerequisites are given, we can make inferences over current and future labels as well as future feature sequences. 

% key contribution 
In this paper, we adress all of the problems in Table \ref{tab:actions} simultaneously with a generative, temporal latent variable model that can capture the complex dependencies of continuous features as well as discrete labels over time. With real-time deployment in mind, we focus on noisy 3D recordings of human joint positions and object features recorded with Kinect devices. 

% model SVRNN 
In detail, we propose a semi-supervised variational recurrent neural network (SVRNN), as described in Section \ref{sec:SVRNN}, which inherits the generative capacities of a variational autoencoder (VAE) \cite{kingma2013auto, rezende2014stochastic}, extends these to temporal data \cite{chung2015recurrent} and combines them with a discriminative model in a semi-supervised fashion. The semi-supervised VAE \cite{kingma2014semi} can handle labeled and unlabeled data. This property allows us to propagate label information over time even during testing and therefore to generate possible future action and motion sequences. 
In addition, we propose to make use of the hierarchical label structure in human activities in the form of a hierarchical SVRNN (HSVRNN), as described in Section  \ref{sec:hSVRNN} and to model the dependencies between multiple entities, such as a human and objects or two interacting humans, by extending the model to a multi-entity SVRNN (ME-SVRNN), as introduced in Section  \ref{sec:meSVRNN}.  

% experiments
We benchmark our model on the Cornell Activity Dataset 120 (CAD -120) \cite{koppula2013learning}, the UTKinect-Action3D Dataset \cite{xia2012view} and the Stony Brook University Kinect Interaction Dataset (SBU) \cite{kiwon_hau3d12}. We find that our mixed-data,  multi-task approach outperforms or performs comparably to state-of-the-art, task-specific methods in the different tasks listed in Table \ref{tab:actions} (see Section \ref{sec:experiments}).   

The contributions of this paper are 1) the development of a semi-supervised variational RNN which can infer and propagate semantic and continuous information over time and therefore allows for online multi-task deployment, 2) extensions of the model to capture hierarchical and multimodal data, 3) a unification of six common tasks in human activity modeling into a single problem statement and experimental baselines for future work.
\section{Background}
\label{sec:background}

Our approach builds on three basic ingredients, namely Variational Autoencoders, or VAEs, (Section \ref{sec:vae}), their semi-supervised equivalent (Section \ref{sec:SVAE}) and a recurrent version (Section \ref{sec:rvae}). To ease understanding of later sections, we will here introduce each of these concepts and reference to relevant literature for further details. First, we will introduce the notation used in this paper. 

\noindent \textbf{Notation}
We represent continuous data points by $\mathbf{x}$, discrete labels by $\mathbf{y}$ and $\mathbf{c}$ and latent variables by  $\mathbf{z}$. The hidden state of a recurrent neural network (RNN) unit at time $t$ is denoted by $h_t$. Similarly, time-dependent random variables are indexed by $t$, e.g. $\mathbf{x_t}$. Distributions $p_\theta$ commonly depend on parameters $\theta$. For the sake of brevity, we will neglect this dependency in the following discussion. $\{\mathbf{x_t}\}_{t=1:N}$ is a set of data points in the interval 1 to $N$. $(\mathbf{a},\mathbf{b})$ denotes a pair, $(\mathbf{a} \cap \mathbf{b})$ denotes an intersection and $[\mathbf{a},\mathbf{b}]$ a concatenation of variables $\mathbf{a}$ and $\mathbf{b}$.

% ---------------------------------------------------------------------------
% VAE
\subsection{Variational autoencoders}
\label{sec:vae}

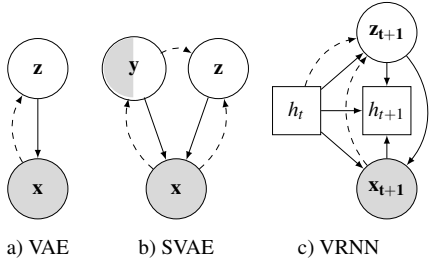
\begin{figure}[t!]

\centering   
\scalebox{0.8}{\pgfdeclarelayer{background}
\pgfdeclarelayer{foreground}
\pgfsetlayers{background,main,foreground}

\begin{tikzpicture}

\centering

\tikzstyle{observed} = [circle,draw=black, fill=gray!30,  minimum size=1cm,  inner sep=1.5pt]
\tikzstyle{unobserved} = [circle,draw=black, fill=white!30,  minimum size=1cm,  inner sep=1.5pt]

\tikzstyle{semiobserved} = [circle,draw=black, fill=white!30,pattern=north east lines, pattern color=gray,  minimum size=1cm,  inner sep=1.5pt]

\tikzstyle{empty} = [circle,draw=black, fill=white!0,minimum size=1.06cm,  inner sep=1.5pt]

\tikzstyle{rnnnode} = [rectangle,draw=black, fill=white!30,  minimum size=0.8cm,  inner sep=1.5pt]

\tikzstyle{textnode} = [rectangle,draw=white, fill=white!30,  minimum size=0.8cm,  inner sep=1.5pt]

\tikzstyle{condition} = [ellipse,draw=black, fill=green!30,    inner sep=2.5pt]
\tikzstyle{action} = [rectangle,draw=black, fill=yellow!30,   inner sep=2.5pt]
\tikzstyle{control} = [regular polygon,regular polygon sides=4, draw, fill=white!11,  text badly centered,  text width=1.0em,  inner sep=1.5pt]
\tikzstyle{arrowline} = [draw,color=black, -latex]
\tikzstyle{arrowlined} = [draw,  color=black, -latex , out=5  ]
\tikzstyle{arrowdashed} = [draw,dashed, color=black, -latex  ]
\tikzstyle{arrowdash} = [draw,dashed, color=black, -latex,  bend left=40]
\tikzstyle{arrowbend} = [draw, color=black, -latex, bend right=40]
\tikzstyle{arrowbendleft} = [draw, color=black, -latex, bend left=40]

\tikzstyle{arrowdashbend} = [draw,dashed, color= black, -latex, bend right=15]

\tikzstyle{arrowdashbendleft} = [draw,dashed, color= black, -latex, bend left=30]
\tikzstyle{arrowdashbendleftmore} = [draw,dashed, color= black, -latex, bend left=40]

\tikzstyle{arrowdashbendred} = [draw, dashed,  color= black,, -latex, bend right=35]
\tikzstyle{arrowdashbendless} = [draw, dashed,  color= black,, -latex, bend right=30 ]
 
 % \draw [black,dotted, thick, -latex,  bend left]   (-1.4,-1.5) to (Z);
\tikzstyle{textit} = [draw=none,fill=none]

% ----------------------------------------------------------------------------

\node [observed] at (2, 0.3) (nXt) {$\mathbf{x}$};
\node [unobserved] at (2, 2.3) (nZt) {$\mathbf{z}$};
\node [textnode] at (2, -0.7) (texta) {a) VAE};

\path [arrowline] (nZt) to (nXt); 
\path [arrowdashbendleft] (nXt) to (nZt);

\node [observed] at (4.3, 0.3) (Xt) {$\mathbf{x}$};
\node [unobserved] at (5, 2.3) (Zt) {$\mathbf{z}$};
%\node [semiobserved] at (3.6, 2.3) (Yt) {$\mathbf{Y}$};
\node [empty] at (3.6, 2.3) (ds) { };
 \node[shape=circle,
    draw=white ,  minimum size=1cm,  inner sep=1.5pt,
    circle split part fill={gray!30,white!30}
    ] at (3.6, 2.3) (Yt){$\mathbf{y}$};

\node [textnode] at (4.3, -0.7) (textb) {b) SVAE};
\path [arrowline] (Zt) to (Xt); 
\path [arrowdashbendless] (Xt) to (Zt);

\path [arrowline] (Yt) to (Xt); 
\path [arrowdashbendleft] (Xt) to (Yt);
\path [arrowdashbendleft] (Yt) to (Zt);

% ----------------------------------------------------------------------------

 \node [observed] at (7.8, 0.3) (nnXt1) {$\mathbf{x_{t+1}}$};

 \node [rnnnode] at ( 6.3,  1.6) (nnhzt) {$h_{t}$};

 \node [rnnnode] at (7.8,  1.6) (nnhzt1) {$h_{t+1}$};

\node [unobserved] at ( 7.8, 2.9) (nnCt1) {$\mathbf{z_{t+1}}$};

\node [textnode] at ( 7, -0.7) (textc) {c) VRNN};

 \path [arrowline] (nnhzt) to (nnhzt1); 
 \path [arrowline] (nnhzt) to (nnCt1); 
 \path [arrowline] (nnhzt) to (nnXt1);
 \path [arrowline] (nnCt1) to (nnhzt1);
 \path [arrowline] (nnXt1) to (nnhzt1);
 \path [arrowbendleft] (nnCt1) to (nnXt1);
 \path [arrowdashbendleftmore] (nnXt1) to (nnCt1);
 \path [arrowdashbendleft] (nnhzt) to (nnCt1);

\end{tikzpicture}}
\caption{Model structure of the VAE \textbf{a)}, its semi-supervised version SVAE \textbf{b)}, and the recurrent model VRNN \textbf{c)}. Random variables (circle) and states of RNN hidden units (square) are either observed (gray), unobserved (white) or partially observed (white-gray). The dotted arrows indicate inference connections.}  \vspace{-1.5em}
\label{fig:intromodels}
\end{figure}

Our model builds on VAEs, latent variable models that are combined with an amortized version of variational inference (VI). Amortized VI employs neural networks to learn a function from the data $\mathbf{x}$ to a distribution over the latent variables $q(\mathbf{z}|\mathbf{x})$  that approximates the posterior $p(\mathbf{z}|\mathbf{x})$. Likewise, they learn the likelihood distribution as a function of the latent variables $p(\mathbf{x}|\mathbf{z})$. This mapping is depicted in Figure \ref{fig:intromodels}a). Instead of having to infer $N$ local latent variables for $N$ observed data points, as common in VI, amortized VI requires only the learning of neural network parameters of the functions $q(\mathbf{z}|\mathbf{x})$ and $p(\mathbf{x}|\mathbf{z})$. We call $q(\mathbf{z}|\mathbf{x})$  the recognition network and  $p(\mathbf{x}|\mathbf{z})$ the generative network. 
To sample from a VAE, we first draw a sample from the prior $\mathbf{z} \sim p(\mathbf{z})$ which is then fed to the generative network to yield $\mathbf{x} \sim p(\mathbf{x}|\mathbf{z})$. We refer to \cite{zhang2017advances} for more details. 

% ---------------------------------------------------------------------------
% SVAE
\subsection{Semi-supervised variational autoencoders}
\label{sec:SVAE}
	
To incorporate label information when available, semi-supervised VAEs (SVAE) \cite{kingma2014semi} include a label $\mathbf{y}$ into the generative process $p(\mathbf{x}|\mathbf{z}, \mathbf{y})$ and the recognition network  $q(\mathbf{z}|\mathbf{x}, \mathbf{y})$, as shown in Figure \ref{fig:intromodels}b). To handle unobserved labels, an additional approximate distribution over labels $q(\mathbf{y}|\mathbf{x})$ is learned which can be interpreted as a classifier. When no label is available, the discrete label distribution can be marginalized out, e.g. $q(\mathbf{z}|\mathbf{x}) = \sum_\mathbf{y}q(\mathbf{z}|\mathbf{x}, \mathbf{y})q(\mathbf{y}|\mathbf{x})$. 

\subsection{Recurrent variational autoencoders}
\label{sec:rvae}
% VRNN
VAEs can also be extended to temporal data, so called variational recurrent neural networks (VRNN)  \cite{chung2015recurrent}. Instead of being stationary as in standard VAEs, the prior over the latent variables depends in this case on past observations $p(\mathbf{z_t}|h_{t-1})$, which are encoded in the hidden state of an RNN $h_{t-1}$. Similarly, the approximate distribution $q(\mathbf{z_t}|\mathbf{x_t}, h_{t-1})$ depends on the history as can be seen in Figure \ref{fig:intromodels}c). The advantage of this structure is that data sequences can be generated by sampling from the temporal prior instead of an uninformed prior, i.e. $\mathbf{z_t} \sim p(\mathbf{z_t}|h_{t-1})$. 
\section{Methodology}
\label{sec:method}

Equipped with the background knowledge introduced in the previous section, we will now describe the structure of our proposed model, semi-supervised variational recurrent neural networks (SVRNN), and the inference procedure applied to train them (Section \ref{sec:SVRNN}). We will further detail how to extend the model to capture a hierarchical label structure (Section \ref{sec:hSVRNN}) and to jointly model multiple entities (Section \ref{sec:meSVRNN}).

\subsection{SVRNN}
\label{sec:SVRNN}

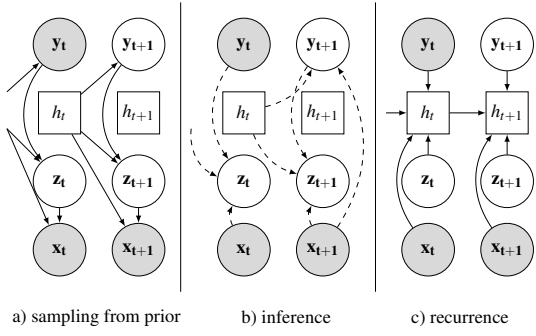
\begin{figure}[t!]
\centering   
\scalebox{0.7}{\pgfdeclarelayer{background}
\pgfdeclarelayer{foreground}
\pgfsetlayers{background,main,foreground}

\begin{tikzpicture}

\centering

\tikzstyle{observed} = [circle,draw=black, fill=gray!30,  minimum size=1cm,  inner sep=1.5pt]
\tikzstyle{unobserved} = [circle,draw=black, fill=white!30,  minimum size=1cm,  inner sep=1.5pt]

\tikzstyle{rnnnode} = [rectangle,draw=black, fill=white!30,  minimum size=0.8cm,  inner sep=1.5pt]

\tikzstyle{condition} = [ellipse,draw=black, fill=green!30,    inner sep=2.5pt]
\tikzstyle{action} = [rectangle,draw=black, fill=yellow!30,   inner sep=2.5pt]
\tikzstyle{control} = [regular polygon,regular polygon sides=4, draw, fill=white!11,  text badly centered,  text width=1.0em,  inner sep=1.5pt]
\tikzstyle{arrowline} = [draw,color=black, -latex]
\tikzstyle{arrowlined} = [draw,  color=black, -latex , out=5  ]
\tikzstyle{arrowdashed} = [draw,dashed, color=black, -latex , out=5  ]
\tikzstyle{arrowdash} = [draw,dashed, color=black, -latex,  bend left=40]
\tikzstyle{arrowbend} = [draw, color=black, -latex, bend right=40]
\tikzstyle{arrowbendleft} = [draw, color=black, -latex, bend left=35]

\tikzstyle{arrowdashbend} = [draw,dashed, color= black, -latex, bend right=25]

\tikzstyle{arrowdashbendleft} = [draw,dashed, color= black, -latex, bend left=25]

\tikzstyle{arrowdashbendred} = [draw, dashed,  color= black,, -latex, bend right=35]
\tikzstyle{arrowdashbendless} = [draw, dashed,  color= black,, -latex, bend right=33 ]
 
 % \draw [black,dotted, thick, -latex,  bend left]   (-1.4,-1.5) to (Z);
\tikzstyle{textit} = [draw=none,fill=none]

\node [observed] at (-4, 0.3) (Xt) {$\mathbf{x_t}$};
\node [observed] at (-2.5, 0.3) (Xt1) {$\mathbf{x_{t+1}}$};

\node [unobserved] at (-4, 1.6) (Zt) {$\mathbf{z_t}$};
\node [unobserved] at (-2.5, 1.6) (Zt1) {$\mathbf{z_{t+1}}$};

\node [rnnnode] at (-4, 2.9) (hzt) {$h_t$};
\node [rnnnode] at (-2.5, 2.9) (hzt1) {$h_{t+1}$};

\node [observed] at (-4, 4.2) (Ct) {$\mathbf{y_t}$};
\node [unobserved] at (-2.5, 4.2) (Ct1) {$\mathbf{y_{t+1}}$};

\node [textit] at (-3.3, -1) (a) {a) sampling from prior};

% recurrent to rv
\path [arrowline] (hzt) to (Zt1); 

\path [arrowline] (hzt) to (Xt1); 
 \path [arrowline] (hzt) to (Ct1);

\path [arrowline] (-5,2.6) to (Zt);
\path [arrowline] (-5,3.3) to (Ct);
\path [arrowline] (-5,2.6) to (Xt);

% rv to rv
\path [arrowline] (Zt) to (Xt); 
\path [arrowline] (Zt1) to (Xt1); 
 
\path [arrowbend] (Ct) to (Zt);
\path [arrowbend] (Ct1) to (Zt1);

%\path [arrowbend] (Ct1) to (Zt1); 

% ----------------------------------------------------------------------------

\node [observed] at (-0.5, 0.3) (nXt) {$\mathbf{x_t}$};
\node [observed] at (1.0, 0.3) (nXt1) {$\mathbf{x_{t+1}}$};

\node [unobserved] at (-0.5, 1.6) (nZt) {$\mathbf{z_t}$};
\node [unobserved] at ( 1, 1.6) (nZt1) {$\mathbf{z_{t+1}}$};

\node [rnnnode] at (-0.5, 2.9) (nhzt) {$h_t$};
\node [rnnnode] at ( 1, 2.9) (nhzt1) {$h_{t+1}$};

\node [observed] at (-0.5, 4.2) (nCt) {$\mathbf{y_t}$};
\node [unobserved] at ( 1, 4.2) (nCt1) {$\mathbf{y_{t+1}}$};

\path [arrowdashbendred] (nCt) to (nZt);  
\path [arrowdashbendred] (nCt1) to (nZt1);

\path [arrowdashbend] (nhzt) to (nZt1); 
\path [arrowdashbend] (nhzt) to (nCt1);

\path [arrowdashbendleft] (nXt) to (nZt);
\path [arrowdashbendleft] (nXt1) to (nZt1);

\path [arrowdashbendless] (nXt1) to (nCt1);

\path [arrowdashbendred] (-1.5,2.6) to (nZt);
 
  \draw (-1.7,-0.5) -- (-1.7,5);
\node [textit] at (0.3, -1) (b) {b) inference};

% ----------------------------------------------------------------------------

\node [observed] at (3, 0.3) (nnXt) {$\mathbf{x_t}$};
\node [observed] at (4.5, 0.3) (nnXt1) {$\mathbf{x_{t+1}}$};

\node [unobserved] at (3, 1.6) (nnZt) {$\mathbf{z_t}$};
\node [unobserved] at ( 4.5, 1.6) (nnZt1) {$\mathbf{z_{t+1}}$};

\node [rnnnode] at (3, 2.9) (nnhzt) {$h_t$};
\node [rnnnode] at ( 4.5, 2.9) (nnhzt1) {$h_{t+1}$};

\node [observed] at (3, 4.2) (nnCt) {$\mathbf{y_t}$};
\node [unobserved] at ( 4.5, 4.2) (nnCt1) {$\mathbf{y_{t+1}}$};

 \draw (2,-0.5) -- (2,5);

 % recurrent
\path [arrowline] (nnhzt) to (nnhzt1);

\path [arrowline] (nnCt) to (nnhzt); 
\path [arrowline] (nnCt1) to (nnhzt1);  

\path [arrowline] (nnZt) to (nnhzt); 
\path [arrowline] (nnZt1) to (nnhzt1);  

\path [arrowline] (2.2 ,2.9) to (nnhzt); 
\path [arrowbendleft] (nnXt) to (nnhzt); 
\path [arrowbendleft] (nnXt1) to (nnhzt1);

 \node [textit] at (3.6, -1) (c) {c) recurrence};

\end{tikzpicture}}
\caption{Information flow through SVRNN. a) Passing samples from the prior through the generative network. b) Information passing through the inference network. c) The recurrent update. Node appearance follows Figure \ref{fig:intromodels}.} 
\label{fig:model}
\end{figure}

In the SVRNN, the model is trained on a dataset with temporal structure $D = \{D^L, D^U\}$ consisting of the set $L$ of labeled time steps $D^L = \{\bmxt,\bmyt\}_{t \in L} \sim \pem(\bmxt,\bmyt)$ and the set $U$ of unlabeled observations $D^U = \{\bmxt\}_{t \in U} \sim  \pem(\bmxt)$. $\pem$ denotes the empirical distribution. Further we assume that the temporal process is governed by latent variables $\bmzt$, whose distribution $p(\mathbf{z_t}|\bmyt, h_{t-1})$ depends on a deterministic function of the history up to time $t$: $h_{t-1} = f(x_{<t}, y_{<t}, z_{<t})$. The generative process is as follows
\begin{align}
\mathbf{y_t} \sim p(\mathbf{y_t}|h_{t-1}), \ \mathbf{z_t} \sim p(\mathbf{z_t}|\mathbf{y_t}, h_{t-1}), \  \mathbf{x_t} \sim p(\mathbf{x_t}|\mathbf{y_t}, \mathbf{z_t}, h_{t-1}), \label{eq:generative}
\end{align} 
where $p(\mathbf{y_t}|h_{t-1})$ and $p(\mathbf{z_t}|\mathbf{y_t}, h_{t-1})$ are time-dependent priors, as shown in Figure \ref{fig:model}a). To fit this model to the dataset at hand, we need to estimate the posterior over the unobserved variables $p(\mathbf{y_t}|\mathbf{x_t}, h_{t-1})$ and $p(\mathbf{z_t}|\mathbf{x_t}, \mathbf{y_t}, h_{t-1})$ which is intractable. Therefore we resign to amortized VI and approximate the posterior with a simpler distribution  $q(\mathbf{y_t}, \mathbf{z_t}|\mathbf{x_t}, h_{t-1}) = q(\mathbf{y_t}|\mathbf{x_t}, h_{t-1})q(\mathbf{z_t}|\mathbf{x_t}, \mathbf{y_t}, h_{t-1})$, as shown in Figure \ref{fig:model}b). To minimize the distance between the approximate and posterior distributions, we optimize the variational lower bound of the marginal likelihood $\mathcal{L}(p(D)) $. As the distribution over $\mathbf{y_t}$ is only required when it is unobserved, the bound decomposes as follows
\begin{align}
& \quad \quad \quad \mathcal{L}  (p(D))  \geq  \mathcal{L}^L  + \mathcal{L}^U  + \alpha\mathcal{T}^L\label{eq:loglikone}\\
-\mathcal{L}^L &=  \sum_{t \in L}  \mathbb{E}_{q(\bmzt|\bmxt, \bmyt, \bmhlt )}[log(p(\bmxt| \bmyt, \bmzt, \bmhlt))] \label{eq:logliklabel}   \\
& \quad  - KL(q(\bmzt|\bmxt, \bmyt , \bmhlt )||p(\bmzt|  \bmyt, \bmhlt ))  + log(p(\bmyt)) \nonumber   \\
 \mathcal{T}^L &=-\sum_{t \in L}\mathbb{E}_{\pem(\bmyt, \bmxt)} log (p(\bmyt|\bmhlt )q(\bmyt|\bmxt, \bmhlt )) \label{eq:loglikH} \\
 -\mathcal{L}^U &=  \sum_{t \in U}  \mathbb{E}_{q(\bmyt, \bmzt|\bmxt, \bmhlt )} \big{[} log(p(\bmxt| \bmyt, \bmzt, \bmhlt ))] \  \label{eq:loglikno}  \\
 &   \quad -   KL(q(\bmzt|\bmxt, \bmyt, \bmhlt  )||p(\bmzt| \bmyt, \bmhlt  ))   \nonumber \\
 &   \quad  -   KL(q(\bmyt|\bmxt, \bmhlt )||p(\bmyt|\bmhlt )).  \nonumber
\end{align}
$\mathcal{L}^L$ and $\mathcal{L}^U$ are the lower bounds for labeled and unlabeled data points respectively, while $\mathcal{T}^L$ is an additional term that encourages $p(\bmyt|\bmhlt)$ and $q(\bmyt|\bmxt, \bmhlt)$ to follow the data distribution over $\mathbf{y_t}$. This lower bound is optimized jointly. We assume the latent variables $\mathbf{z_t}$ to be i.i.d. Gaussian distributed. The categorical distribution  over $\mathbf{y_t}$ is determined by parameters $\pi = \{\pi_i\}_{i=1:N_{class}}$. To model such discrete distributions, we apply the Gumbel trick \cite{jang2016categorical,maddison2016concrete}. The history $h_{t-1}$ is modeled with a Long short-term memory (LSTM) unit \cite{hochreiter1997long}. For more details, we refer the reader to the background work discussed in Section \ref{sec:background} and the Supplementary material.

\subsection{Hierarchical SVRNN}
\label{sec:hSVRNN}

Human activity can often be described by hierarchical semantic labels. For example, the label \textit{cleaning} might be parent to the labels \textit{vacuuming} and \textit{scrubbing}. While we here describe how to model a hierarchy consisting of two label layers, the number of layers is not constrained. Let the parent random variable of $\mathbf{y_t}$ be represented by $\mathbf{c_t}$. 
To incorporate $\mathbf{c_t}$ we extend the model by additional prior and approximate distributions, $p(\mathbf{c_t}|\bmhlt)$ and $q(\mathbf{c_t}|\bmxt, \bmhlt)$. The latent state $\bmzt$ at time $t$ depends on both $\bmyt$ and $\bmct$. Thus, the dependency of $\mathbf{y_t}$ and $\mathbf{z_t}$ on $\mathbf{c_t}$ is modeled by conditioning as follows $q(\mathbf{y_t}|\bmxt, \bmct, \bmhlt ), p(\mathbf{y_t}| \bmct, \bmhlt), q(\mathbf{z_t}|\bmxt, \bmyt, \bmct, \bmhlt)$ and $p(\mathbf{z_t}| \bmyt,  \bmct, \bmhlt)$. 

Instead of partitioning the dataset into two parts, $D = \{D^L, D^U\}$, the additional variable requires us to divide it into four parts, $D = \{D^{L_y,L_c}, D^{L_y,U_c},D^{U_y,L_c},D^{U_y,U_c}\}$, where $D^{L_y,L_c} = \{\bmxt, \bmyt, \bmct\}_{t \in (L_y \cap L_c) } \sim  \pem(\bmxt, \bmyt, \bmct)$, $D^{L_y ,  U_c} = \{\bmxt, \bmyt\}_{t \in (L_y \cap U_c) } \sim  \pem(\bmxt, \bmyt)$, $D^{U_y,L_c} = \{\bmxt,   \bmct\}_{t \in (U_y  \cap L_c) } \sim  \pem(\bmxt, \bmct)$ and $D^{U_y , U_c} = \{\bmxt\}_{t \in (U_y \cap U_c) } \sim  \pem(\bmxt )$. This means that the lower bound in Equation \ref{eq:loglikone} is extended to 
\begin{align}
&  \mathcal{L}  (p(D))  \geq \sum_{l_y, l_c}  \mathcal{L}^{l_y,l_c} 
+ \alpha(\mathcal{T}^{L_y,L_c} + \mathcal{T}^{L_y,U_c} + \mathcal{T}^{U_y,L_c}), \label{eq:logliktwo} 
\end{align}
where $l_y \in \{L_y, U_y\}$ and $l_c \in \{L_c, U_c\}$. The lower bounds $\mathcal{L}^{l_y,l_c}$ and additional terms $\mathcal{T}^{l_y,l_c}$ follow the same structure as Equation \ref{eq:logliklabel}, \ref{eq:loglikH} and \ref{eq:loglikno} and are detailed in the Supplementary material.

\subsection{Multi-entity SVRNN}
\label{sec:meSVRNN}

To model different entities, we allow these to share information between each other over time. The structure and information flow of this model is a design choice. In our case, these entities consist of the human $H$ and $o \in [1, N_o]$ additional entities, such as objects or other humans. We denote the dependency of variables on their source by $(\mathbf{x_t}^H, \mathbf{y_t}^H, \mathbf{z_t}^H, h_t^H)$ and $\{(\mathbf{x_t}^o, \mathbf{y_t}^o, \mathbf{z_t}^o, h_t^o)\}_{o \in 1:N_o}$. Further, we summarize the history and current observation of all additional entities by $h_t^O=\sum_{o}h_t^o$ and  $\mathbf{x_t^O}=\sum_{o}\mathbf{x_t^o}$ respectively. Instead of only conditioning on its own history and observation, as described in Section \ref{sec:SVRNN}, we let the entities share information by conditioning on others' history and observations. Specifically, the model of the human receives information from all additional entities, while these receive information from the human model. 
Let $\mathbf{x_t^{AB}} = [\mathbf{x^A_{t}}, \mathbf{x_{t}^B}]$ and $h_t^{AB} = [h^A_{t}, h_{t}^B]$ for $A,B \in (H,O,o)$.
The structure of the prior and approximate distribution then become  
$p(\mathbf{\mathbf{y^H_t}}|h^{HO}_{t-1})$, $p(\mathbf{\mathbf{z^H_t}}|\mathbf{y^H_t}, h^{HO}_{t-1})$,
$q(\mathbf{\mathbf{y^H_t}}|\mathbf{x^{HO}_t}, h_{t-1}^{HO})$ and $q(\mathbf{\mathbf{z^H_t}}|\mathbf{x^{HO}_t}, \mathbf{y^H_t}, h_{t-1}^{HO})$ for the human, and $p(\mathbf{\mathbf{y^o_t}}|h^{oH}_{t-1})$, $p(\mathbf{z^o_t}|\mathbf{y^o_t}, h^{oH}_{t-1})$, $q(\mathbf{\mathbf{y^o_t}}|\mathbf{x^{oH}_t}, h^{oH}_{t-1})$ and $q(\mathbf{z^o_t}| \mathbf{x^{oH}_t}, \mathbf{y^o_t}, h_{t-1}^{oH})$ for each additional entity $o \in  1:N_o$, We assume that the labels for all entities are observed and unobserved at the same points in time. Therefore, the lower bound in Equation \ref{eq:loglikone} is only extended by summing over all entities:
\begin{align}
\mathcal{L}  (p(D))  \geq \sum_{e \in \{H, \ o \in [1, N_o]\}} \mathcal{L}^{L_e}  + \mathcal{L}^{U_e}  + \alpha\mathcal{T}^{L_e}, \label{eq:loglikthree}
\end{align}
where $\mathcal{L}^{L_e}, \mathcal{L}^{U_e}$ and $\mathcal{T}^{L_e}$ depend on the probability distributions associated with entity $e$ and take the same form as in Equation \ref{eq:loglikone}. This model can be extended to a hierarchical version ME-HSVRNN.  

\subsection{Classify, predict, detect, anticipate and generate}
\label{sec:methods}

Once trained, we make use of the different components of our model to solve the problems listed in Table \ref{tab:actions}. We describe only the procedures for the SVRNN as the other models follow the same ideas.  \newline
\noindent  \textbf{Classify, predict and detect actions:} To classify or detect at time $t$, we choose the largest of the weights $\pi^{q_y} = \{\pi^{q_y}_i\}_{i=1:N_{class}}$ of the categorical distribution  $q(\mathbf{y_t}|\bmxt, \bmhlt)$. Classification is performed at the end of the sequence, while prediction and detection are performed at all time steps.  \newline
\noindent \textbf{Anticipate actions:} To anticipate a label after time $t$, we make use of the prior, which does not depend on the current observation $\bmxt$. Thus, for time $t+1$, we choose the largest of the weights $\pi^{p_y} = \{\pi^{p_y}_i\}_{i=1:N_{class}}$ of the categorical distributions $p(\mathbf{y_t}|\bmhlt)$. To anticipate several steps into the future, we need to generate both future observations and future labels with help of the priors $p(\mathbf{y_t}|\bmhlt)$ and  $p(\mathbf{z_t}|\bmyt, \bmhlt)$ as described below. \newline
\noindent \textbf{Predict and generate motion:}
To sample an observation sequence $\{\mathbf{x_t},\mathbf{y_t}\}_{t>t'}$ after time $t'$, we follow the generative process in Equation \ref{eq:generative} for each $t$ by propagating the sampled observations and generating with help of the approximate distribution $\mathbf{y_t} \sim q(\mathbf{y_t}|\bmxt, h_{t-1}), \ \mathbf{z_t} \sim q(\mathbf{z_t}|\bmxt, \mathbf{y_t}, h_{t-1}),\ \mathbf{x_t} \sim p(\mathbf{x_t}|\mathbf{y_t}, \mathbf{z_t}, h_{t-1})$ for each $t$. This method is used to predict a sequence, by averaging over several samples of the distributions.   
\section{Related work}
\label{sec:related}

Before presenting the experimental results, we will point to relevant prior work both when it comes to methodology (Section \ref{sec:relatedmodel}) and to human action classification (Section \ref{sec:relatedactions0}), action detection and prediction (Section \ref{sec:relatedactions1}), action anticipation (Section \ref{sec:relatedactions2}) and human motion prediction and synthesis (Section \ref{sec:relatedmotions}). As each of these fields is rich in literature, we will concentrate on a few, highly related works that consider 3D skeletal recordings. 

\subsection{Recurrent latent variable models with class information}
\label{sec:relatedmodel}

Recurrent latent variable models that encode explicit semantic information have mostly been developed in the natural language processing community. The aim of \cite{XuSDT17} is sequence classification. They encode a whole sequence into a single latent variable, while static class information, such as sentiment, that lasts over a whole sequence is modeled in a semi-supervised fashion.  
A similar model is suggested in \cite{zhou2017} for sequence transduction. Multiple semantic labels, such as part of speech or tense, are encoded into a control signal $y$. 
Sequence transduction is also the topic of \cite{NaradowskyCMW18}. In contrast to \cite{zhou2017}, the latent space is assumed to resemble a morphological structure, i.e., that at every word in a sentence is assigned latent lemmata and morphological tags. 
While this discrete structure is optimal for language, continuous variables, such as trajectories, require continuous latent dynamics. These are modeled by \cite{yingzhen2018disentangled}, who divide the latent space into static (e.g. appearance) and dynamic (e.g. trajectory) variables which are approximated in an unsupervised fashion. While this model lends itself to sequence generation, it is not able to incorporate explicit semantic information. 
In contrast to \cite{XuSDT17,yingzhen2018disentangled} and \cite{zhou2017}, our model incorporates semantic information that changes over the cause of the sequence, such as composable action sequences, and does simultaneously model continuous dynamics.

\subsection{Human activity classification}
\label{sec:relatedactions0}

3D human action classification is a broad field which has been covered by several surveys, e.g \cite{berretti2018representation} and \cite{wang2018rgb}. Traditionally, the problem of classifying a motion sequence has been a two-stage process of feature extraction followed by time series modeling, e.g. with Hidden Markov Models \cite{yamato1992recognizing}. Developments in deep learning have led to fusing these two steps. Both convolutional neural networks, e.g. \cite{du2015skeleton, kim2017interpretable}, and recurrent neural network architectures, e.g. \cite{du2015hierarchical, liu2017skeleton}, have been adapted to this task. Recent developments include the explicit modeling of co-occurrences between joints \cite{zhu2016co} and the introduction of attention mechanisms that focus on action-relevant joints, the so called Global Context-Aware Attention LSTM (GCA-LSTM) \cite{liu2018skeleton}. A different approach are View Adaptive LSTMs (VA-LSTMs) which learn to transform the skeleton spatially to facilitate classification \cite{zhang2017view}. Compared to these approaches, we adopt a semi-supervised, probabilistic latent variable model which is not fine-tuned to the type of input data.

\subsection{Human activity prediction and detection}
\label{sec:relatedactions1}

Activity prediction and detection are related in the sense that both methods require classification before the whole sequence has been observed. Detection, however, aims also at determining the onset of an action within a data stream. 
To encourage early recognition, \cite{aliakbarian2017encouraging} defines a loss that penalizes immediate mistakes more than long-term false classifications. A more adaptive approach is proposed in \cite{liu2018ssnet}, namely a convolutional neural network with different scales which are automatically selected such that actions can be predicted early on. 
In order to detect action onsets, \cite{gupta2016scale} combines class-specific pose templates with dynamic time warping. Similarly, such pose templates are used by the authors of 
\cite{lillo2016hierarchical} who couple these with variables describing actions at different levels in a hierarchical model. Instead of templates, 
\cite{li2016online} introduces an LSTM that is trained to both classify and predict the onset and offset of an action. In contrast to these approaches, we propose a generative, semi-supervised model, which proposes action hypotheses from the first frame and onward. As we do not constrain the temporal dynamics of the distribution over labels, the model learns to detect action changes online.

\subsection{Human activity anticipation}
\label{sec:relatedactions2}
Activity anticipation aims at predicting semantic labels of actions that have not yet been initiated. This spatio-temporal problem has been addressed with anticipatory temporal conditional random fields (ATCRF) \cite{koppula2016anticipating1}, which augment conditional random fields (CRFs) with predictive, temporal components. In a more recent work, structural RNNs (S-RNNs) have been used to classify and predict activity and affordance labels by modeling the edges and nodes of CRFs with RNNs \cite{jain2015structural}. Instead of a supervised approach, our semi-supervised generative model propagates label information over time and anticipates the label of the next action by definition.   

\subsection{Human motion prediction and synthesis}
\label{sec:relatedmotions}

Recent advances in human motion prediction are based on deep neural networks. As an example, S-RNNs have been adapted to model the dependencies between limbs as nodes of a graphical model \cite{jain2015structural}. However, RNN based models have been outperformed by a window-based representation learning method \cite{butepage17} and 
suffer among others from an initial prediction error and propagated errors \cite{martinez2017human}. When the network has to predict residuals, or velocity, in an unsupervised fashion (residual unsupervised, RU) these problems can be overcome \cite{martinez2017human}.
Human motion modeling with generative models has previously been approached with Restricted Bolzmann Machines \cite{taylor2006modeling}, Gaussian Processes \cite{wang2008gaussian} and Variational Auroencoders \cite{butepage2017anticipating}. In \cite{habibie2017recurrent}, a recurrent variational autoencoder is developed to synthesize human motion with a control signal. Our model differs in several aspects from this approach as we explicitly learn a generative model over both observations and labels and  make use of time-dependent priors. 

\section{Experiments}
\label{sec:experiments}

\begin{table}[b!]     
\caption{\label{tab:actionclass}Average F1 score for activity (Act), sub-action (SAct) and object affordances (Aff) for detection and anticipation (CAD-120).}
\begin{tabular}{|c|c|c|c|c|c|}
\hline
 & \multicolumn{2}{|c|}{Detection}  & \multicolumn{2}{|c|}{Anticipation}  & \multicolumn{1}{|c|}{ }  \\
\hline
Method &  SAct  & Aff & SAct  & Aff & Act\\ \hline
ATCRF \cite{koppula2016anticipating1}  &  86.4  & 85.2 &  40.6 & 41.4 & 94.1 \\ \hline
S-RNN \cite{jain2015structural} & 83.2 & 91.1  & 65.6 & 80.9 & - \\ \hline
S-RNN (SF) & 69.6  & 84.8 & 53.9 & 74.3 & - \\ \hline
SVRNN & 83.4  & 88.3 & 67.7 & 81.4 & -\\ \hline
ME-SVRNN &  89.8   &   90.5  &   77.1  &   82.1  & -\\ \hline
ME-HSVRNN & \textbf{90.1}  &  \textbf{91.2} &  \textbf{79.9} &  \textbf{83.2} & \textbf{96.0} \\ \hline
\end{tabular}

\vspace{-0.3cm}
\end{table} 

In this section, we describe both experimental design and results. 
First, we detail the datasets (Section \ref{sec:datasets}). In the following, we investigate the ability of our model solve multiple tasks: to detect and anticipate human activity (Section \ref{sec:actiondetection}), to detect and predict actions (Section \ref{sec:actionprediction}) and to classify actions (\ref{sec:actionclassification}). 
The final experiments center around the prediction and synthesis of continuous human motion (Section \ref{sec:motionprediction}). Model structure and training procedures are detailed in the Supplementary Material.

Note that while we present results on individual tasks for the sake of comparison with other methods, our approach solves the remaining tasks simultaneously. Thus, when we present results for e.g. sequence classification, the trained model can also be used for e.g. action detection or motion prediction.

% ------------------ Datasets and experimental settings --------------------------
\subsection{Datasets}
\label{sec:datasets}
 
 We apply our models to the Cornell Activity Dataset 120 (CAD -120) \cite{koppula2013learning}, the UTKinect-Action3D Dataset (UTK) \cite{xia2012view} and the Stony Brook University Kinect Interaction Dataset (SBU) \cite{kiwon_hau3d12}.

% ------------------ Datasets CAD --------------------------
\noindent \textbf{CAD-120:} 
The CAD-120 dataset \cite{koppula2013learning} consists of 4 subjects performing 10 high-level tasks, such as \textit{cleaning a microwave} or \textit{having a meal}, in 3 trials each. These activities are further annotated with 10 sub-actions, such as \textit{moving} and \textit{eating} and 12 object affordances, such as \textit{movable} and \textit{openable}. In this work we focus on detecting and anticipating the activities, sub-actions and affordances. 
Our results rely on four-fold cross-validation with the same folds as used in \cite{koppula2013learning}. For comparison, we train S-RNN models, for which code is provided online, on these four folds and under the same conditions as described in \cite{jain2015structural}. We use the features extracted in \cite{koppula2013learning} and pre-process these as in \cite{jain2015structural}. The object models share all parameters, i.e., we effectively learn one human model and one object model both in the single- and multi-entity case.

% ------------------ Datasets UT --------------------------
\noindent \textbf{UTKinect-Action3D Dataset:} 
The UTKinect-Action3D Dataset (UTK) \cite{xia2012view} consists of 10 subjects each recorded twice performing 10 actions in a row. The sequences are recorded with a kinect device (30 fps) and the extracted skeletons consist of 20 joints. Due to  high inter-subject,  intra-class  and view-point variations, this dataset is challenging. While most previous work has used the segmented action sequences for action classification, we are aiming at action detection and prediction, i.e., the model has to detect action onset and classify the actions correctly. This is demanding as the longest recording contains 1388 frames that need to be categorized. The actions in each recording do not immediately follow each other but are disrupted by long periods of unlabeled frames. As our model is semi-supervised, these unobserved data labels can be  incorporated naturally and do not require the introduction of e.g. an additional $unknown$ label class.
We train our model on five subjects and test on the remaining five subjects.

% ------------------ Datasets SBU --------------------------

 \begin{table}[b!]  
\vspace{-0.2cm}
\caption{\label{tab:UTpredict} F1 score for action prediction with history (with H) and without history (without H) on the UTK dataset.}
\vspace{-0.3cm}
\center
\begin{tabular}{|c|c|c|c|c| }
\hline 
Observed &  25 \% & 50 \% & 75 \% & 100 \% \\ \hline
CT \cite{gupta2016scale} & -  &  - & - & 81.8 \\ \hline 
SVRNN (unseg) &  61.0  &  78.0 & 84.0 & \textbf{89.0} \\ \hline  
SVRNN (seg) &  29.0  &  48.0 & 67.0 & 74.0 \\ \hline 
\end{tabular}
\end{table}

\begin{table*}[t!]  
\vspace{-0.5cm}
\centering
\parbox{.35\linewidth}{
\centering
\caption{\label{tab:sbuclass} Average accuracy for interactive sequence classification (SBU). Note that the all results, except ours, were produced with methods highly tuned towards sequence classification.}
\begin{tabular}{|c|c| }
\hline 
Method &  Acc \% \\ \hline
Joint Feat. \cite{yun2012two} &  80.3   \\ \hline
Joint Feat. \cite{ji2014interactive} &  86.9   \\ \hline  
Co-occ. RNN \cite{zhu2016co} &  90.4   \\ \hline  
\textbf{ME-SVRNN}   &  91.0   \\ \hline 
STA-LSTM \cite{song2017end} &  91.5   \\ \hline 
GCA-LSTM  \cite{liu2018skeleton} &  94.9   \\ \hline
VA-LSTM \cite{zhang2017view} &  \textbf{97.2}   \\ \hline
\end{tabular}
}
\hfill
\centering
\parbox{.63\linewidth}{
\centering
\caption{\label{tab:sbupred} Average motion prediction error for interactive sequences (SBU). Note that the RU method focuses its computational resources on motion prediction, while our model is regularized by its probabilistic formulation and the need to infer the class label. }
\begin{tabular}{|c||c|c|c|c||c|c|c|c| }
\hline 
\multicolumn{1}{|c||}{Method} & \multicolumn{4}{|c||}{RU \cite{martinez2017human}} & \multicolumn{4}{|c|}{ME-SVRNN}   \\ \hline
Time (ms) &   260 & 400 & 530  & 660  &  260  & 400  & 530  & 660 \\ \hline
approach & \textbf{0.10} & \textbf{0.22} & \textbf{0.37} & \textbf{0.47} &  0.17  & 0.37 & 0.55 & 0.72 \\ \hline
punch & \textbf{0.29}& 0.69& 1.25& 1.60 &  0.34& \textbf{0.63} & \textbf{0.81} & \textbf{1.00} \\ \hline
hug &  0.31 & 0.75& 1.37& 1.76& \textbf{0.30} & \textbf{0.61} & \textbf{0.80}  & \textbf{1.00} \\ \hline
push &  0.29& 0.66& 1.15& 1.42& \textbf{0.19} & \textbf{0.35} & \textbf{0.45} & \textbf{0.56} \\ \hline
kick & \textbf{0.20} & \textbf{0.50} & 0.91& 1.15 &  0.37& 0.71& \textbf{0.90} & \textbf{1.14} \\ \hline
\end{tabular}

}
\vspace{-0.4cm}
\end{table*}

\noindent \textbf{SBU Kinect Interaction Dataset:} 
The SBU dataset \cite{kiwon_hau3d12} contains around 300 recordings of seven actors (21 pairs of two actors) performing eight different interactive activities such as \textit{hugging}, \textit{pushing} and \textit{shaking hands}. The data was collected with a kinect device at 15 fps. While the dataset contains color and depth image, we make use of the 3D coordinates of 15 joints of each subject. As these measurements are very noisy, we smooth the joint locations over time \cite{du2015hierarchical}. We follow the five-fold cross-validation suggested by \cite{kiwon_hau3d12}, which splits the dataset into five folds of four to five actors. 
On the basis of the SBU dataset, we investigate sequence classification as well as prediction and generation of interactive human motion over the range of around 660 ms (10 frames). 
In order to model two distinct entities, we assign the two actors in each recording the label \textit{active} or \textit{passive}. For example, during the action \textit{kicking} the active subject kicks while the \textit{passive} subject avoids the kick. In a more equal interaction such as \textit{shaking hands}, the \textit{active} actor is the one who initiates the action. We list these labels for all recorded sequences in the supplementary material.

\begin{figure}[b]
 \vspace{-1.5em}
\center
\includegraphics[width=0.45\textwidth]{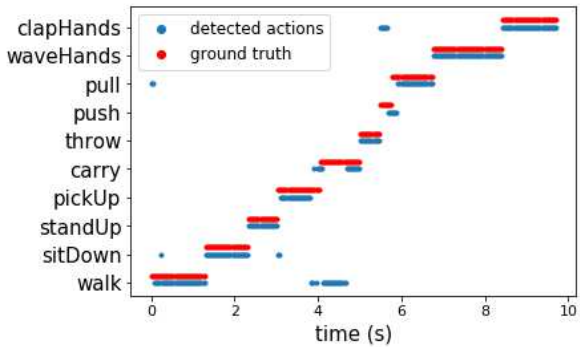}
\caption{\label{fig:utdetect} The detected and ground truth actions of a single test recording from the UTK dataset over time. We only display the labeled frames of the test sequence.}
\end{figure}

\subsection{Activity detection and anticipation}
\label{sec:actiondetection}

In this section, we focus on the capabilities of our models to detect and anticipate semantic activity labels. We present experimental results on the inference of actions as well as sub-actions and affordance labels based on the CAD-120 dataset.

\noindent \textbf{CAD-120:}
Following related work \cite{jain2015structural, koppula2013learning}, we investigate the detection and anticipation performance of our model for sub-actions (SAct), object affordances (Aff) and high-level actions (Act). Detection entails classification of the current observation at time $t$ and anticipation measures the predictive performance for the next observation at time $t+1$. 

In Table \ref{tab:actionclass} we present the results for the baseline models ATCRF \cite{koppula2016anticipating1}  and S-RNN (as reported in \cite{jain2015structural} and reproduced on the same folds (SF) as we use here). We compare these to the performance of the vanilla SVRNN, the multi-entity ME-SVRNN and a multi-entity hierarchical ME-HSVRNN. We see that especially the anticipation of sub-actions gains in performance when incorporating information from the object entities (ME-SVRNN). Further improvements are achieved when the hierarchical label structure is included (ME-HSVRNN).

  \begin{figure*}[th!]
 \centering
\includegraphics[width=0.8\textwidth]{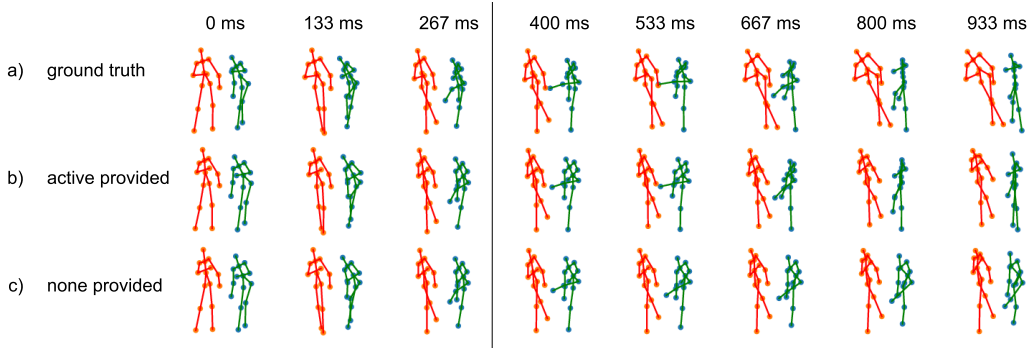}
\caption{\label{fig:predgen} Visualization of a $kicking$ action of the active (green) and passive (red) subject. The line indicates when the model starts to generate. We show \textbf{a)} the ground truth, \textbf{b)} generating the passive subject for 530 ms and reconstructing the active subject and  \textbf{c)} generating both subjects for 530 ms.}
 \vspace{-1.5em}
\end{figure*}

\subsection{Action detection and prediction}
\label{sec:actionprediction}

In this section, we focus on the capabilities of our models to detect and predict semantic labels.  We test the performance of our model on the UTK dataset.

\noindent \textbf{UTKinect-Action3D Dataset:}
As far as we are aware, only one comparable work, based on class templates \cite{gupta2016scale}, has attempted to detect actions on the UTK dataset. \cite{lillo2016hierarchical} only reports results on jointly detecting which actions are performed and which body parts are used (F1 score=69.0).  We assume action $a$ to be detected if the majority of observations within the ground truth time interval are inferred to belong to action $a$.
We compare the F1 score averaged over all classes after having observed 100 \% of each action to \cite{gupta2016scale} in Table \ref{tab:UTpredict}. More generally, we see that the model is able to detect actions with only a short or no delay. This is apparent when we measure the F1 score for partially observed action sequences, namely when the model has observed 25 \%, 50 \%, 75 \% or 100 \% of the current action in Table \ref{tab:UTpredict}. We present results for action detection in context of the previous actions, i.e., on the unsegmented sequence (unseg), and for action prediction based only on the current action segment (seg). On average, this corresponds to having observed 8, 16, 25 or 33 frames of the ongoing action. 
As listed in Table \ref{tab:UTpredict}, the F1 score increases continuously the more of the action has been observed. At 75 \% the SVRNN outperforms the results reported in \cite{gupta2016scale} which are based on 100 \% of the action interval. When segmented, the performance is lower as our model has not been trained to predict actions without history.

Further, we visualize the detected and ground truth action sequence of one unsegmented test sample in Figure \ref{fig:utdetect} and in form of a video 
%in the supplementary material.
here https://www.youtube.com/watch?v=XfgztgOhuCk. 
In this test sequence, the action $carry$ is partially confused with $walking$ which might be caused by the lack of meta-data such as that the subject is holding an object.

\subsection{Action classification}
\label{sec:actionclassification}

Action classification aims at determining the action class at the end of an observed motion sequence. We apply this method to classify interactive actions of two actors (SBU).

\noindent \textbf{SBU Kinect Interaction Dataset:}
To classify a sequence, we average over the last three time steps of the sequence. The classification accuracy of our model is compared to state-of-the-art models in Table \ref{tab:sbuclass}. Our model achieves comparable performance to most of the related work. It needs to be kept in mind that the other models are task-specific and data-dependent and are not able to e.g. predict labels or human motion. Thus, the computational resources of the other models are solely directed towards classification.

\subsection{Motion prediction and synthesis}
\label{sec:motionprediction}

Finally, we present results on feature prediction and synthesis. We present results on the SBU dataset, which means that we predict and generate two interacting subjects. We present additional results on the Human3.6M dataset \cite{h36m_pami} (H36M) in the Supplementary material. The H36M dataset consists of motion capture data and is often used for human motion prediction experiments. 

\noindent \textbf{SBU Kinect Interaction Dataset:}
We compare the predictive performance to a state-of-the-art human motion prediction model (RU) \cite{martinez2017human}. This model learns the residuals (velocity) in an unsupervised fashion and is provided with a one-hot vector indicating the class. To be comparable, we also model the residuals. Thus, the main differences between the RU and the ME-SVRNN are that we a) formulate a probabilistic, latent variable model, b) combine information of both subjects and c) model an explicit belief over the class distribution. To compare, we let both models predict ten frames given the first six frames of the actions $approaching$, $punching$, $hugging$,  $pushing$ and $kicking$. The error is computed as the accumulated squared distance between the ground truth and the prediction of both subjects up to frame $t$.  We present the results for 260, 400, 530 and 660 ms in Table \ref{tab:sbupred}. While the RU outperforms our model for $approaching$ and some measurements at +260 ms, the ME-SVRNN performs better during long-term predictions.    

In addition to prediction, our generative model allows us to sample possible future trajectories. In the case of multiple entities, we can either generate all entity sequences or provide the observation sequence of one entity while generating the other. In Figure \ref{fig:predgen} we present samples of the action $kicking$. The upper row shows the ground truth. The middle row was produced by providing the model with the sequence of the $active$ subject while generating the sequence of the $passive$ subject. In the lower row, the sequences of both subjects are generated. A video of additional results on the UTKinect-Action3D Dataset, showcasing the inference of a discrete class change and a sample of the joint trajectories following this class change, can be found 
%in the supplementary material.
here https://www.youtube.com/watch?v=EoOz5aqpWtk.

\section{Conclusion}
\label{sec:discussion}

Human activity modeling poses a number of challenging spatio-temporal problems. In this work we proposed a semi-supervised generative model which learns to represent semantic labels and continuous feature observations over time. In this way, the model is able to simultaneously classify, predict, detect and anticipate discrete labels and to predict and generate feature sequences. When extended to model multiple entities and hierarchical label structures, our approach is able to tackle complex human activity sequences. While most previous work has been centered around task-specific modeling, we suggest that joint modeling of continuous observations and semantic information, whenever available, forces the model to learn a more holistic representation which can be used to solve many different tasks. In future work, we plan to extend our model to more challenging semantic information such as raw text and to incorporate multiple modalities.

{\small
\bibliographystyle{ieee}
\bibliography{egpaper_final}
}

\section{Supplementary material}
 
This is the supplementary material of the paper \textit{A Probabilistic Semi-Supervised Approach to Multi-Task Human Activity Modeling}. Here we describe details of the derivation of the semi-supervised variational recurrent neural network (SVRNN) in Section \ref{sec:app:SVRNN} and its hierarchical version in Section \ref{sec:app:hier}. Furthermore, we describe the network architectures and data processing steps for all experiments in Section \ref{sec:app:network} and \ref{sec:app:data} respectively. We list the labels of active and passive subjects for the Stony Brook University Kinect Interaction Dataset (SBU) \cite{kiwon_hau3d12} in Section \ref{sec:app:labels}.   Finally, we present additional results on human motion prediction in Section \ref{sec:app:h36m}.

\subsection{SVRNN}
\label{sec:app:SVRNN}

The derivation of the lower bound follows the description in \cite{kingma2014semi} with a number of exceptions. 
First of all, we assume that the approximate distribution factorizes $q(\bmyt, \bmzt |\bmxt, \bmhlt) = q(\bmyt|\bmxt, \bmhlt)q(\bmzt |\bmxt, \bmyt, \bmhlt)$ and the prior over the latent variable $\mathbf{z_t}$ does depend on the label and the history $p(\bmzt|\bmyt, \bmhlt)$.

Secondly, we use two different priors on the discrete random variable $\mathbf{\mathbf{y_t}}$ depending on whether the data point has been observed $t \in L$ or is unobserved $t \in U$. We apply a uniform prior $p(\bmyt)$ for $\forall t \in L$ and a history-dependent prior $p(\bmyt|\bmhlt )$ for $\forall t \in U$, which follows the Gumbel-Softmax distribution.

Finally, as the prior on the discrete variables $\mathbf{\mathbf{y_t}}$ is history-dependent, we want to encourage it to encode the information provided in the labeled data points. Therefore,   we add not only an additional term for the approximate distribution $q(\bmyt|\bmxt, \bmhlt)$ but also for the prior distribution $p(\bmyt|\bmhlt)$.

\subsection{Hierarchical SVRNN}
\label{sec:app:hier}

The derivation of Equation 6 follows the discussion in Section \ref{sec:app:SVRNN} and in \cite{kingma2014semi}. Below, we detail the specific form of each component.  For the sake of brevity, we introduce the following notation
\begin{align}
p(\mathbf{x_t}|b^x) &:= p(\bmxt|\bmyt,\bmct, \bmzt, \bmhlt) \nonumber \\
p(\mathbf{z_t}|b^z) &:= p(\bmzt|\bmyt,\bmct, \bmhlt) \nonumber \\
p(\mathbf{y_t}|b^y) &:= p(\bmyt|\bmct, \bmhlt) \nonumber \\
p(\mathbf{c_t}|b^c) &:= p(\bmct|\bmhlt) \nonumber \\
q(\mathbf{z_t}|b^z) &:= q(\bmzt|\bmxt, \bmyt,\bmct, \bmhlt) \nonumber \\
q(\mathbf{y_t}|b^y) &:= q(\bmyt|\bmxt,\bmct, \bmhlt) \nonumber \\
q(\mathbf{c_t}|b^c) &:= q(\bmct|\bmxt,\bmhlt), \nonumber 
\end{align}
where $b^e$ denotes the conditional (background) variables of variable $e$.
When both labels are present $t \in (L_y \cap L_c)$ the lower bound and additional term take the following form
\begin{align}
-  \mathcal{L}^{L_y, L_c}  &=  \sum_{t}  \mathbb{E}_{q(\bmzt|b^z})[log(p(\bmxt|b^x))]   + log(p(\bmyt)) \nonumber   \\
& \quad  - KL(q(\bmzt|b^z)||p(\bmzt|b^z))  + log(p(\bmct))  \nonumber \\
 \mathcal{T}^{L_y, L_c} &=-\sum_{t}\mathbb{E}_{\pem(\bmyt, \bmct, \bmxt)} log (p(\bmyt| b^y))q(\bmyt|b^y)) \nonumber  \\
 & \quad -\sum_{t }\mathbb{E}_{\pem(\bmct,\bmxt)} log (p(\bmct|b^c)q(\bmct|b^c)). \nonumber 
\end{align}

When only the label $\mathbf{\mathbf{y_t}}$ has been observed $t \in (L_y \cap U_c)$, the lower bound and additional term take the following form
\begin{align}
-   \mathcal{L}^{L_y, U_c}  & =  \sum_{t} \mathbb{E}_{q(\bmct|b^c)} \Big{[} \mathbb{E}_{q(\bmzt|b^z)}[log(p(\bmxt|b^x))]  \nonumber \\ 
& \quad - KL(q(\bmzt|b^z)||p(\bmzt| b^z )) \Big{]} + log(p(\bmyt)) \nonumber\\
& \quad -  KL(q(\bmct|b^c )||p(\bmct|b^c ))   \nonumber \\
 \mathcal{T}^{L_y, U_c} & = -\sum_{t } \mathbb{E}_{q(\bmct|b^c )} \Big{[} \mathbb{E}_{\pem(\bmyt, \bmxt)} log (p(\bmyt|b^y)) \nonumber \\ 
  & \quad + \sum_{t }\mathbb{E}_{\pem(\bmyt, \bmxt)} log (q(\bmyt|b^y )) \Big{]}.  \nonumber
\end{align}

When only the label $\mathbf{\mathbf{c_t}}$ has been observed $t \in (U_y \cap L_c)$, the lower bound and additional term take the following form
\begin{align}
-  \mathcal{L}^{U_y, L_c} & =  \sum_{t} \mathbb{E}_{q(\bmyt|b^y)} \big{[} \mathbb{E}_{q(\bmzt|b^z)}[log(p(\bmxt|b^x))] \  \nonumber \\
 & \quad  -   KL(q(\bmzt|b^z)||p(\bmzt|b^z))\big{]} + log(p(\bmct)) \nonumber \\
 & \quad  -   KL(q(\bmyt|b^y)||p(\bmyt|b^y))  \nonumber \\ 
  \mathcal{T}^{U_y, L_c} & =-\sum_{t}\mathbb{E}_{\pem(\bmct,\bmxt)} log (p(\bmct|b^c)q(\bmct|b^c)). \nonumber
\end{align}

When only no label has been observed $t \in (U_y \cap U_c)$, the lower bound takes the following form
\begin{align}
-  \mathcal{L}^{U_y, U_c} & = \sum_{t} \mathbb{E}_{q(\bmct|b^c)} \Big{[} \mathbb{E}_{q(\bmyt|b^y)} \big{[}   \mathbb{E}_{q(\bmzt|b^z)}[log(p(\bmxt|b^x))] \  \nonumber \\
 & \quad  -   KL(q(\bmzt|b^z)||p(\bmzt|b^z))\big{]} \nonumber \\
 & \quad  -   KL(q(\bmyt|b^y)||p(\bmyt|b^y))\Big{]}  \nonumber \\
 & \quad  -  KL(q(\bmct|b^c)||p(\bmct|b^c)). \nonumber
\end{align}

\subsection{Network architecture}
\label{sec:app:network}

In this section, we begin by describing the overall structure and follow up with details on the specific number of units for each experiment. 

We represent the unobserved labels as a stochastic vector and the observed labels as a one-hot vector.
The distributions over labels are given by fully connected neural networks with a Gumbel-Softmax output layer. 
The input is given by a concatenation $[\mathbf{x_t}, h_{t-1}]$ for the approximate label distribution and by $h_{t-1}$ for the prior label distribution. In case of a hierarchical structure, we concatenate even the parent label, e.g. $[\mathbf{x_t},\mathbf{c_t}, h_{t-1}]$ for $\mathbf{y_t}$.

The distributions over latent variables $\mathbf{\mathbf{z_t}}$ are given by fully connected neural networks that output the parameters of a Gaussian $(\mu_t, \sigma_t)$. The input is given by $[\mathbf{x_t}, \mathbf{y_t}, h_{t-1}]$ for the approximate distribution in the case of SVRNN and $[\mathbf{x_t}, \mathbf{y_t},\mathbf{c_t}, h_{t-1}]$  in the case of HSVRNN  and by $[\mathbf{y_t},\mathbf{c_t}, h_{t-1}]$ for the prior distribution. When a label has not been observed, we propagate a sample from the respective Gumbel-Softmax distribution. 

The recurrent unit receives the input $[\mathbf{x_t}, \mathbf{y_t}, \mathbf{z_t}, h_{t-1}]$ in the case of SVRNN and $[\mathbf{x_t}, \mathbf{y_t},\mathbf{c_t}, \mathbf{z_t}, h_{t-1}]$ in the case of HSVRNN.

Fully connected neural networks are used to reconstruct the next observation based on the input  $[\mathbf{x_t}, \mathbf{y_t}, \mathbf{z_t}]$ in the case of SVRNN and $[\mathbf{x_t}, \mathbf{y_t},\mathbf{c_t}, \mathbf{z_t}]$ in the case of HSVRNN.

When multiple entities are combined, the same structure as discussed above is used. However, in this case the observations and history features are concatenated $\mathbf{x_t} = \mathbf{x_t}^{AB} = [\mathbf{x^A_{t}}, \mathbf{x_{t}^B}]$ and $h_t = h_t^{AB} = [h^A_{t}, h_{t}^B]$ for $A,B \in (H,O,o)$ for the respective entities. 

We use the $tanh$ non-linearity for all layers except for the output and latent variables layers. The recurrent layers consist of LSTM units. 

\paragraph{CAD-120 - action detection and anticipation}

We always map the input to 256 dimensions for each entity with a fully connected layer.
As each entity follows the same pattern, the details below do not distinguish between them.
The approximate distribution and prior over $\bmyt$ of dimension $dim_y$ is given by $input-256-dim_y$.
The approximate distribution and prior over $\bmct$ of dimension $dim_c$ is given by $input-256-dim_c$.
The approximate distribution and prior over $\bmzt$ of dimension $dim_z=256$ are given by $input-256-256-dim_z$.
The size of the hidden state of the recurrent layer is $256$. 
The reconstruction of the observation $\bmxt$ of dimension $dim_x$ is given by $input-512-512-dim_x$.

\paragraph{UTKinect-Action3D - action detection and prediction}

We always map the input to 516 dimensions for each entity with a fully connected layer.
The approximate distribution and prior over $\bmyt$ of dimension $dim_y$ is given by $input-516-dim_y$.
The approximate distribution and prior over $\bmct$ of dimension $dim_c$ is given by $input-516-dim_c$.
The approximate distribution and prior over $\bmzt$ of dimension $dim_z=516$ are given by $input-516-516-dim_z$.
The size of the hidden state of the recurrent layers is $516$. We have three layers.
The reconstruction of the observation $\bmxt$ of dimension $dim_x$ is given by $input-1032-1032-dim_x$.

\paragraph{SBU - action classification and motion generation and prediction}

We always map the input to 516 dimensions for each entity with a fully connected layer.
As each entity follows the same pattern, the details below do not distinguish between them.
The approximate distribution and prior over $\bmyt$ of dimension $dim_y$ is given by $input-516-dim_y$.
The approximate distribution and prior over $\bmct$ of dimension $dim_c$ is given by $input-516-dim_c$.
The approximate distribution and prior over $\bmzt$ of dimension $dim_z=516$ are given by $input-516-516-dim_z$.
The size of the hidden state of the recurrent layers is $516$. We have three layers.
The reconstruction of the observation $\bmxt$ of dimension $dim_x$ is given by $input-1032-1032-dim_x$.
The same model is used for the additional experiments presented in Section \ref{sec:app:h36m}.

\begin{table*}[t!]     
 
\centering
\begin{tabular}{|c||c|c|c|c||c|c|c|c| }
\hline 
\multicolumn{1}{|c||}{Method} & \multicolumn{4}{|c||}{RU \cite{martinez2017human}} & \multicolumn{4}{|c|}{ME-SVRNN}   \\ \hline
Time (ms) &   80 & 160 & 320  & 400  &  80  & 160  & 320  & 400 \\ \hline
walking  & \textbf{0.37} & 0.67 & 1.01 & 1.12                & 0.43  & \textbf{0.57} & \textbf{0.76} & \textbf{0.87} \\ \hline
eating  & \textbf{0.29} & 0.52 & 0.9 & 1.10                 &  0.36 & \textbf{0.42} & \textbf{0.65} & \textbf{0.81} \\ \hline
smoking  & 0.37 & 0.68 & 1.22 & 1.36                        & \textbf{0.36} & \textbf{0.57} & \textbf{0.97} & \textbf{0.98}  \\ \hline
discussion & \textbf{0.36} & \textbf{0.76} & 1.14 & 1.28                & 0.52 &  0.81 & \textbf{1.03} & \textbf{1.09} \\ \hline
directions  & \textbf{0.47} & \textbf{0.77} & 1.05 & 1.23           & 0.59 & 0.79 & \textbf{0.86} & \textbf{0.95} \\ \hline
greeting  &    \textbf{0.56} & \textbf{0.96} & \textbf{1.47} & 1.66    & 0.77 & 1.08 & \textbf{1.47} & \textbf{1.62} \\ \hline
phoning  &\textbf{ 0.68} & \textbf{1.22} & 1.76 & 1.96               &  0.71 & \textbf{1.22} & \textbf{1.58} & \textbf{1.70}  \\ \hline
posing  &  1.04 & 1.31 & 1.97 & 2.26                       & \textbf{0.46} & \textbf{0.69} & \textbf{1.3} & \textbf{1.58}  \\ \hline
purchases  & \textbf{0.65} & \textbf{0.92} & 1.32 & 1.44             & 0.81 &1.01 & \textbf{1.26} & \textbf{1.33}  \\ \hline
sitting  &  \textbf{0.53} & 0.91 & 1.56 & 1.85                & 0.55& \textbf{0.81} & \textbf{1.15} & \textbf{1.29}  \\ \hline
sitting down  &     0.53 & 0.99 & 1.55 & 1.80                 & \textbf{0.51} & \textbf{0.86} & \textbf{1.23} & \textbf{1.36} \\ \hline
taking photo   & \textbf{0.34} & 0.68 & 1.17 & 1.38                &  0.36 & \textbf{0.63} & \textbf{0.91} & \textbf{1.03}  \\ \hline
waiting  & \textbf{0.40} & 0.78 & 1.34 & 1.61               &  0.45 & \textbf{0.72} & \textbf{1.19} & \textbf{1.39} \\ \hline
walking dog  & \textbf{0.58} & \textbf{0.99} & 1.39 & 1.58        &  0.72 &  0.96 & \textbf{1.27} & \textbf{1.44}  \\ \hline
walking together  & \textbf{0.35} & 0.70 & 1.11 & 1.26         & 0.43  & \textbf{0.67} & \textbf{0.84} & \textbf{0.91}  \\ \hline
\end{tabular}
\caption{\label{tab:h36m} Average motion prediction error for motion capture sequences (H36M).}

\end{table*}

\subsection{Data preprocessing  and training}
\label{sec:app:data}

We always set $\alpha$ equal to the number of all features and the temperature parameter of the Gumbel-Softmax distribution to $0.1$.
For all fully connected layers except the output layers and parameters of the latent variables, we apply a dropout rate of 0.1.

\paragraph{CAD-120 - action detection and anticipation}

We use the features extracted in \cite{koppula2013learning} and preprocess these as in \cite{jain2015structural}. 
The features contain information about the human subject and the objects, which are modeled jointly in the multi-entity condition. The hierarchical structure is given by the high-level task governing the sub-activities.
The features extracted by \cite{koppula2013learning} assign a sub-activity label and affordance labels to the human subject and the objects respectively at each time step. In each batch, we mark ca. 25 \% of the labels as unobserved. 
We apply a learning rate of $0.001$ and cut the gradients at an absolute value of $5$. The results are averaged over 20 evaluations of our probabilistic models.

\paragraph{UTKinect-Action3D - action detection and prediction}

The skeletons are centered around the root joint to reduce variability between recordings. All unlabeled observations between two labeled action intervals are set to be unobserved.  In each training epoch, 10 \% of the remaining action labels are randomly assigned to be unobserved.  We apply a learning rate of $0.001$ and cut the gradients at an absolute value of $5$. The results are averaged over 10 evaluations of our probabilistic model.  

\paragraph{SBU - action classification}

For classification, the model is provided with the whole sequence as evidence and is trained to predict a single frame at each time step. To force the network to encode the sequence label over a long time period, we label only the last seven frames of each recording with the respective interaction label and assume the remaining  labels to be unobserved.
We apply a learning rate of $0.001$ and cut the gradients at an absolute value of $5$. 

\paragraph{SBU - motion generation and synthesis}
 
For sequence prediction, we provide the model with six frames of observations from which it needs to predict the ten following frames. In this case, we label only the last three frames of each data point in the mini-batch. The remaining labels are assumed to be unobserved. The results are averaged over 20 evaluations of our probabilistic model.
We apply a learning rate of $0.0005$ and cut the gradients at an absolute value of $5$.

\subsection{Human motion prediction - H36M}
\label{sec:app:h36m}

In this section we present additional results for human motion prediction on the Human3.6M dataset \cite{h36m_pami} (H36M).
This motion capture dataset consists of seven actors performing 15 actions in two trials each. We follow the data processing and error computation as described in \cite{martinez2017human}. Both the residual unsupervised model (RU) \cite{martinez2017human} and our SVRNN model are trained to predict 10 frames given the last 6 frames. The network architecture and modeling approach of the SVRNN  follow the model used in the \textit{SBU - motion generation and synthesis} experiments. The resulting errors for different actions in the range of 80, 160, 320 and 400 ms are listed in Table \ref{tab:h36m}. The SVRNN outperforms the RU model especially in long-term ( $>$ 80 ms) predictions.

\subsection{SBU active and passive labeling}
\label{sec:app:labels}

Below we list the labeling of active and passive subjects in the SBU dataset \cite{kiwon_hau3d12}. Each recording sequence is labeled by action class, subject id and activity level (active or passive) as follows $recordingId;actionLabel;subjectId;activityLevel$, e.g. $s03s06;08;002;0$. We use $0$ for active and $1$ for passive. During actions with distinct roles, such as $kicking$, the assignment of $active$ and $passive$ is straight forward. Actions such as $shaking \ hands$ offer a less clear labeling. We aimed at labeling the action-initiating actor as $active$ in these cases.
 
\noindent \textbf{Fold 1}

\noindent
s01s02;01;001;1
s01s02;01;002;0
s01s02;02;001;1
s01s02;02;002;0
s01s02;03;001;1
s01s02;03;002;0
s01s02;04;001;1
s01s02;04;002;0
s01s02;05;001;1
s01s02;06;001;1
s01s02;07;001;1
s01s02;07;002;0
s01s02;07;003;0
s01s02;08;001;1
s01s02;08;002;0
s03s04;01;001;0
s03s04;01;002;1
s03s04;02;001;0
s03s04;02;002;1
s03s04;03;001;0
s03s04;03;002;1
s03s04;04;001;0
s03s04;04;002;1
s03s04;05;001;0
s03s04;06;001;0
s03s04;07;001;0
s03s04;07;002;1
s03s04;08;001;0
s03s04;08;002;1
s05s02;01;001;1
s05s02;01;002;0
s05s02;02;001;1
s05s02;02;002;0
s05s02;03;001;1
s05s02;03;002;0
s05s02;04;001;1
s05s02;04;002;0
s05s02;04;003;1
s05s02;06;001;1
s05s02;08;001;1
s05s02;08;002;0
s06s04;01;001;0
s06s04;01;002;1
s06s04;02;001;0
s06s04;02;002;1
s06s04;03;001;0
s06s04;03;002;1
s06s04;04;001;0
s06s04;04;002;1
s06s04;05;001;0
s06s04;06;001;0
s06s04;07;001;0
s06s04;07;002;1
s06s04;08;001;0
s06s04;08;002;1

\noindent \textbf{Fold 2}

\noindent
s02s03;01;001;0
s02s03;01;002;1
s02s03;02;001;0
s02s03;02;002;1
s02s03;03;001;0
s02s03;03;002;1
s02s03;04;001;0
s02s03;04;002;1
s02s03;06;001;0
s02s03;07;001;0
s02s03;07;002;1
s02s03;08;001;1
s02s07;01;001;0
s02s07;01;002;1
s02s07;02;001;0
s02s07;02;002;1
s02s07;03;001;0
s02s07;03;002;0
s02s07;03;003;1
s02s07;04;001;0
s02s07;04;002;1
s02s07;05;001;0
s02s07;06;001;0
s02s07;07;001;0
s02s07;07;002;1
s02s07;08;001;0
s02s07;08;002;1
s03s05;01;001;1
s03s05;02;001;1
s03s05;02;002;0
s03s05;03;001;1
s03s05;03;002;0
s03s05;04;001;1
s03s05;04;002;0
s03s05;05;001;0
s03s05;06;001;1
s03s05;07;001;0
s03s05;07;002;1
s03s05;08;001;0
s03s05;08;002;1
s05s03;01;001;0
s05s03;01;002;1
s05s03;02;001;0
s05s03;02;002;1
s05s03;03;001;0
s05s03;04;001;0
s05s03;04;002;1
s05s03;05;001;0
s05s03;06;001;0
s05s03;07;001;0
s05s03;08;001;0
s05s03;08;002;1

\noindent \textbf{Fold 3}

\noindent
s01s03;01;001;0
s01s03;01;002;1
s01s03;02;001;0
s01s03;02;002;1
s01s03;03;001;0
s01s03;03;002;1
s01s03;04;001;0
s01s03;04;002;1
s01s03;05;001;0
s01s03;06;001;0
s01s03;07;001;0
s01s03;08;001;0
s01s03;08;002;1
s01s03;08;003;1
s01s07;01;001;1
s01s07;01;002;0
s01s07;02;001;1
s01s07;02;002;0
s01s07;03;001;1
s01s07;03;002;0
s01s07;04;001;1
s01s07;04;002;0
s01s07;05;001;1
s01s07;06;001;1
s01s07;07;001;1
s01s07;07;002;0
s01s07;08;001;1
s01s07;08;002;0
s07s01;01;001;0
s07s01;01;002;1
s07s01;02;001;0
s07s01;02;002;1
s07s01;03;001;0
s07s01;03;002;1
s07s01;04;001;0
s07s01;04;002;1
s07s01;05;001;0
s07s01;06;001;1
s07s01;07;001;0
s07s01;07;002;1
s07s01;08;001;0
s07s01;08;002;1
s07s03;01;001;1
s07s03;01;002;0
s07s03;02;001;1
s07s03;02;002;0
s07s03;03;001;1
s07s03;03;002;0
s07s03;04;001;1
s07s03;04;002;0
s07s03;05;001;1
s07s03;06;001;1
s07s03;07;001;1
s07s03;07;002;0
s07s03;08;001;1
s07s03;08;002;0

\noindent \textbf{Fold 4}

\noindent
s02s01;01;001;0
s02s01;01;002;1
s02s01;01;003;1
s02s01;02;001;0
s02s01;02;002;1
s02s01;02;003;1
s02s01;03;001;0
s02s01;03;002;1
s02s01;03;003;0
s02s01;04;001;0
s02s01;05;001;0
s02s01;06;001;0
s02s01;07;001;0
s02s01;07;002;1
s02s01;08;001;0
s02s06;01;001;0
s02s06;01;002;1
s02s06;02;001;0
s02s06;02;002;1
s02s06;03;001;0
s02s06;03;002;1
s02s06;04;001;0
s02s06;04;002;1
s02s06;05;001;0
s02s06;06;001;0
s02s06;07;001;0
s02s06;07;002;1
s02s06;08;001;0
s02s06;08;002;1
s03s02;01;001;0
s03s02;01;002;1
s03s02;02;001;0
s03s02;02;002;1
s03s02;03;001;0
s03s02;03;002;1
s03s02;04;001;1
s03s02;05;001;0
s03s02;06;001;1
s03s02;07;001;0
s03s02;07;002;1
s03s02;08;001;0
s03s02;08;002;1
s03s06;01;001;1
s03s06;01;002;0
s03s06;02;001;1
s03s06;02;002;0
s03s06;03;001;0
s03s06;04;001;1
s03s06;04;002;0
s03s06;06;001;0
s03s06;07;001;1
s03s06;07;002;0
s03s06;08;001;1
s03s06;08;002;0

\noindent \textbf{Fold 5}

\noindent
s04s03;01;001;0
s04s03;01;002;1
s04s03;02;001;0
s04s03;02;002;1
s04s03;03;001;0
s04s03;03;002;1
s04s03;04;001;0
s04s03;04;002;1
s04s03;05;001;1
s04s03;06;001;0
s04s03;07;001;0
s04s03;07;002;1
s04s03;08;001;0
s04s03;08;002;1
s04s06;01;001;0
s04s06;01;002;1
s04s06;02;001;0
s04s06;02;002;1
s04s06;03;001;0
s04s06;03;002;1
s04s06;04;001;0
s04s06;04;002;1
s04s06;05;001;0
s04s06;06;001;1
s04s06;07;001;0
s04s06;07;002;1
s04s06;08;001;0
s04s06;08;002;1
s06s02;01;001;1
s06s02;01;002;0
s06s02;02;001;1
s06s02;02;002;0
s06s02;03;001;0
s06s02;04;001;1
s06s02;04;002;0
s06s02;05;001;1
s06s02;06;001;1
s06s02;07;001;0
s06s02;07;002;1
s06s02;08;001;0
s06s03;01;001;1
s06s03;01;002;0
s06s03;02;001;1
s06s03;02;002;0
s06s03;03;001;1
s06s03;03;002;0
s06s03;04;001;1
s06s03;04;002;0
s06s03;05;001;1
s06s03;06;001;0
s06s03;07;001;1
s06s03;07;002;1

\end{document}